\documentclass{article}

\usepackage{arxiv}

\usepackage[utf8]{inputenc} 
\usepackage[T1]{fontenc}    
\usepackage{hyperref}       
\usepackage{url}            
\usepackage{booktabs}       
\usepackage{amsfonts}       
\usepackage{nicefrac}       
\usepackage{microtype}      
\usepackage{graphicx}
\usepackage{natbib}
\usepackage{doi}
\usepackage{verbatim}
\usepackage{caption}
\usepackage{subcaption}

\title{TAD: A Large-Scale Benchmark for Traffic Accidents Detection from Video Surveillance}

\author{ \href{https://orcid.org/0000-0001-9496-7607}{\includegraphics[scale=0.06]{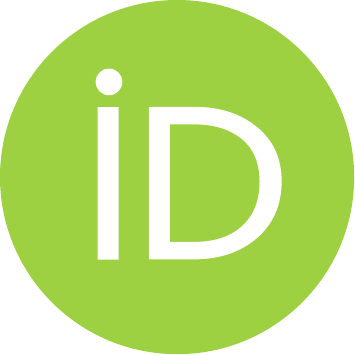}\hspace{1mm}Yajun Xu \thanks{These authors contributed equally to this maniscript} \thanks{ AI Innovation and Application Center of China
Unicom, Beijing, China}}, Chuwen Huang \footnotemark[1] \footnotemark[2], Yibing Nan  \thanks{Corresponding authors (ccpcnanan@126.com, sg\_lian@163.com)} \footnotemark[2], Shiguo Lian \footnotemark[3] \footnotemark[2]}

\hypersetup{
pdftitle={A template for the arxiv style},
pdfsubject={q-bio.NC, q-bio.QM},
pdfauthor={David S.~Hippocampus, Elias D.~Striatum},
pdfkeywords={First keyword, Second keyword, More},
}

\begin{document}
\maketitle

\begin{abstract}

Automatic traffic accidents detection has appealed to the machine vision community due to its implications on the development of autonomous intelligent transportation systems (ITS) and importance to traffic safety. Most previous studies on efficient analysis and prediction of traffic accidents, however, have used small-scale datasets with limited coverage, which limits their effect and applicability. Existing datasets in traffic accidents are either small-scale, not from surveillance cameras, not open-sourced, or not built for freeway scenes. Since accidents happened in freeways tend to cause serious damage and are too fast to catch the spot. An open-sourced datasets targeting on freeway traffic accidents collected from surveillance cameras is in great need and of practical importance. In order to help the vision community address these shortcomings, we endeavor to collect video data of real traffic accidents that covered abundant scenes. After integration and annotation by various dimensions, a large-scale traffic accidents dataset named TAD is proposed in this work. Various experiments on image classification, object detection, and video classification tasks, using public mainstream vision algorithms or frameworks are conducted in this work to demonstrate performance of different methods. The proposed dataset together with the experimental results are presented as a new benchmark to improve computer vision research, especially in ITS.

\end{abstract}


\section{Introduction}
Over the past decades, automatic traffic incidents detection, especially traffic accidents detection, has become a prominent subject in machine vision and pattern recognition because of its tremendous application potential in developing autonomous intelligent transportation systems (ITS). Road traffic accidents are among the life threatening issues facing rural as well as urban community. Accidents detection is of great importance for building ITS. According to the statistics from National Statistics Bureau, the death rate of accidents per 10,000 vehicles is 1.57, still posing threats to peoples’ lives. However, accurate identification of traffic accidents is conducive to avoid secondary accidents which will effectively improve the traffic safety. In light of these applications, researchers in academia and industry are gearing efforts to detect accidents early and potentially reduce their damaging consequences. A significant amount of research has been conducted in detecting traffic accidents through visual information. Despite their usefulness, those targeting traffic accidents are facing several challenges in data as follows:

First, lack of enough background visual information of accidents.  Videos from dashboard cameras have been widely deployed in multiple studies (\cite{2019DADA,9712446,9020591,2019A,2020Traffic}) to provide information covering intervals before and after the accidents in the  first-person perspective. However, despite their usefulness in addressing deficiency in data tailored for accidents detection, they remain limited in providing describing environmental conditions about the involved vehicles, which is important for industrial application in accidents detection. With surveillance cameras installed locating at a certain distance from the ground, video surveillance could provide more comprehensive features of traffic conditions such as positions of surrounding vehicles, traffic flow and even the weather, etc. Especially when more than two vehicles are involved in an accident, non-surveillance datasets would be obviously deficient in analyzing causes and recognizing victims. With the increasing popularity of Close Circuit Television (CCTV) in real life for monitoring traffic conditions, datasets of third-person perspective are in greater need for capability of better catering to current and future studies in developing smart traffic governance. 

Second, small scale specified in accidents limiting generalization. With the rapid advancement in deep learning (DL), algorithms are evolved to achieve efficient feature learning while requiring high quality of data. Traditional studies have categorized as one of traffic anomalies. The University of Central Florida Crime (UCF-Crime) dataset (\cite{2018Real}) collected 151 traffic incident surveillance videos from Iowa during 2016-2017. Although it being massive in terms of all anomaly types, contains a rather small proportion of data specified in traffic incidents, which could not satisfy accuracy of accidents detection. Besides, datasets focusing on traffic accidents are less compared to normal activities, as facing resource-constraint problems. 

Third, single type of accidents classification. With the continuous development of traffic system as well as increasing car parc, categories of accidents vary and keep growing. Existing datasets of accidents detection (\cite{Naphade19AIC19,8639160,2017,9190697}), though large-scale show singularity of types of accidents. For example, AI city competition (2019 track 3) (\cite{Naphade19AIC19}) give specific perspective on vehicle anomalies while only focus on stalled vehicles recognition. In fact, the various accidents could be broadly classified in terms of vehicle number or subjects involved, such as rollovers and collisions are often identified as one positive category while demonstrating distinct features of accidents. Accurate and trustworthy collection of such data play an important role in decision making for smart ITS applications, especially for those industry applications relying on surveillance cameras.

Fourth, limited scenes of occurred accidents. Currently, open-sourced data of traffic accidents mainly present as part of anomalies. However, researches (\cite{8639160,9190697}) in this field often collect data from certain scenes such as urban roads or highways. Other types of roads including countryside and expressway are scarcely retained in relative works, while those are worth noting. Single scene restrains model generalization, thus impeding further development in accidents detection.

To address the aforementioned problems in quantity and quality, we present a new dataset named TAD with specific annotations on traffic accidents, which serves as a benchmark to evaluate the robustness and applicability of traffic accident detection algorithms. The dataset is publicly available at \url{https://github.com/yajunbaby/TAD-benchmark}. The contributions of this work can be summarized as below:

\emph{A novel dataset with largest scale and richest types of accidents in the industry.} The dataset we introduce in this paper consists of four types of accidents covering several scenes in real life, especially with the highway scene, which, to our knowledge, is the largest-scale open-sourced dataset focusing on various types of accidents detection. 

\emph{A benchmark based on experiments of mainstream algorithms.} In this study, our goal is to provide a vision dataset which caters to traffic accidents detection of the industry to build ITS. To achieve this purpose, we organize experiments based on three mainstream visual algorithms to ensure its practicality, including image classification, video classification and object detection. The experimental evaluation indicates TAD’s capability to serve as a benchmark and possible applications in the future.

The rest of our paper is organized as follows: we first review related studies focusing on detection of traffic accidents in Section \ref{sec:Related Work}. In Section \ref{sec:Traffic Accidents Dataset}, we show that TAD is a large-scale, specified visual database with quality-controlled images and annotations, and details on dataset construction such as data source, structure, type and standard of annotation are all provided. In Section \ref{sec:Traffic Events Benchmark}, we present several experiments on TAD as application examples. Our goal is to show that TAD can serve as a useful resource for visual recognition. Finally, the future application as well as discussion is provided in Section \ref{sec:Discussion} and \ref{sec:Future Work}.

\section{Related Work}
\label{sec:Related Work}

\subsection{Datasets of traffic accidents}

\subsubsection{Comma-separated values (CSV) format}
There are several datasets organized in CSV format providing information on traffic events, such as OpenStreetMap, US Accident (\cite{2019A}), etc. Datasets of this type usually consist of a variety of intrinsic and contextual attributes of traffic events such as time, latitude and longitude of an incident and sensor, casualties, vehicle type, road type, etc. So according to the indicators provided, the objective of these datasets is more inclined to prediction than detection. Therefore, apart from the giving indicators, it is often offered with information on traffic flow, weather, period-of-day, and points-of-interest.

\subsubsection{Contextual format}
We can find that in the field of traffic detection, the linguistic text description (\cite{2017Identifying}) belongs to the secondary processing of the information of the accident scene, so there is inevitably a corresponding information loss. Moreover, descriptions in dataset vary from researchers to researchers, which may bring about the problem that different texts may differ greatly in description of an identical accident scene, and thus the accident scenes obtained through textual reconstruction may also be different. For a certain dataset, such differences are likely to cause many problems: e.g., inconsistent data caliber and relatively poor migratory capability of the dataset application. Also considering these limitations, despite the current rapid development of model iterations in natural language processing (NLP), there are few works and public detection results on accident detection using NLP models.

\subsubsection{Vision format}
Video data can provide more three-dimensional and direct information than the previously mentioned CSV and text-based datasets. Obviously, video data directly presents the accident scene visually, and the existing vision-based models are well able to support corresponding image feature extraction work. Given a certain number of annotated training data, the common features of traffic accidents can be modeled by designing appropriate neural networks, then the accidents could be detected with relatively little effort. 

Video itself has its own temporal property, and for each individual frame can provide spatial information of the captured scene, it also has a spatial-temporal property, and this property is very helpful to achieve effective traffic prediction. By using vision-based data,  \cite{9523042} developed a detection system with YOLOv5 and decision trees to recognize traffic anomaly events. Therefore, based on such data features, in order to be able to make full use of the feature information of the video, existing algorithms (\cite{2016A,DBLP:conf/cvpr/ZhaoYPZZSZ19}) usually use multi-dimensional features to extract information from the video etc. Many algorithms (\cite{2019Computer,Habib_2021,Jian-feng_2017}) use the underlying model framework with these features to achieve better prediction results than the above CSV and contextual data.

Currently, video-based traffic accident datasets could be divided into mainly two categories: monitored and non-monitored perspective datasets. Visual datasets not collected from surveillance cameras are mostly derived from data provided by automatic driving recorders (\cite{2019DADA}). \cite{2017} introduced a dataset from dashboard cameras respectively, while the former provided crowd-sourced videos, the latter labeled some 1,500 video traffic accidents collected from YouTube. \cite{9190697} presents a new anomaly detection dataset, the highway traffic anomaly (HTA) dataset in order to detect anomalous traffic patterns from dash cam videos of vehicles on highways. Therefore, these datasets are found limited in wider application due to being difficult to migrate to other scenarios of traffic accident detection, except autonomous driving itself.

Datasets from surveillance view can be classified according to the detailed surveillance scenes. Usually, they can be divided into general surveillance scenarios and freeway surveillance scenarios. The former often includes residential streets, campus, etc. These studies are usually dominated by traffic anomalies datasets, such as UCF-Crime video dataset and AI city Competition dataset. Anomalies defined in the former dataset have a wide range of classes, mostly including crimes committed in streets such as shooting, stealing, vandalism, and robbery. Some of them also includes pedestrian. However, traffic accidents are often not included or simply classified as one of the anomalies types, leaving specific information on vehicle anomalies less provided. Since there is a close relationship between the type of traffic accident and the detected scenes, what occurs in the general street and violent crimes such as home invasion and robbery are hardly seen in the scope of abnormal traffic accident detection in highway scenes. If irrelevant data like criminal acts are removed, the amount of data left for accidents detection will not be rich enough.

In summary, it can be seen that the data collected from video surveillance could provide a more macro, comprehensive and objective perspective of traffic accidents detection, improving urban governance and the construction of intelligent transportation system applications. With the extensive use of surveillance cameras in public places, computer vision-based scene understanding has gained a lot of popularity amongst the CV research community. Visual data contains rich information compared to other information sources such as GPS, mobile location, radar signals. Thus, it can play a vital role in detecting or predicting congestion, accidents and other anomalies apart from collecting statistical information about the status of road traffic. Moreover, open-sourced dataset for traffic accident monitoring in high-speed scenarios is even more important, which is the basis of the dataset proposed in this paper. Considering advantages of surveillance video to capture a variety of realistic anomalies, the dataset we proposed is comprised by videos and images from surveillance cameras installed on various scenes, providing different information from the non-surveillance paradigm like first-person view.

\subsection{Detection algorithms of traffic accidents}

The main purpose of our proposed dataset is to provide a benchmark and help improve research on traffic accidents detection, accordingly this section will mainly discuss and compare the previous work on traffic events detection algorithms. There are three major algorithm frameworks of traffic events detection, including classification methods, causal inference methods, and object detection methods.

\subsubsection{Classification algorithm}
In this field, researchers usually choose to develop a supervised learning method which works as a binary classifier distinguishing between images containing damaged vehicles as positive class and images not containing them as negative class such as \cite{2016A} .

As the research progressed, there are some works focusing on the positive sample with accidents and carried out the next level of classification refinement, such as classification based on the severity of the accident occurrence and the resulting classification analysis of accidents (\cite{Jian-feng_2017}). As for methods deployed in this task, statistical methods both parametrical and non-parametrical can be found in recent studies (\cite{2017Identifying}). Based on the rules of statistics and data mining methods for classification decisions, models widely used are support vector machine (SVM, \cite{1995Support}), Random Forest (RF), Multilayer Perceptron (MLP, \cite{1998Artificial}), Naive Bayes (\cite{Webb2010}), etc. As deep learning (DL) method develops, DL technics have been proved effective in visual classification tasks (\cite{8832160,2019Exploring,2020Video,2020Pre}). Models based on Convolutional Neural Network (CNN, \cite{1989Backpropagation}) such as Convolution 3-dimensional (C3D, \cite{7410867}) perform efficiently in traffic anomaly detection on surveillance videos by distinguishing positive and negative cases of traffic accidents. Methods of time series (\cite{2021Automatic,2016Recurrent}) based classification models such as Recursive Neural Network (RNN, \cite{1991Recursive}), Long Short-Term Memory (LSTM, \cite{1997Long}) also reach high accuracy and recall rate.

\subsubsection{Causal inference algorithm}
Some researchers have also found that there is a close relationship between occurrence of traffic accidents and driver behaviors as well as driving environment before the accidents. Studies in this field helps to early anticipation of traffic accidents. By proposing definitions on risks influencing drivers’ behaviors, \cite{2020Who} introduce a framework via casual inference and demonstrate favorable performance  on Honda Research Institute Driving Dataset. \cite{2022Road} develop a dynamic spatial-temporal attention network using videos from dashboard cameras.

\subsubsection{Object detection algorithm}
Object detection is one of the most important computer vision tasks that deals with detecting instances of visual objects of certain classes (such as humans, animals, or cars) in digital images.

As for the algorithms used in the object detection, from histogram of oriented gradient (HOG) Detector to deformable parts model (DPM) and then to CNN-based models, it has achieved rapid development. Among them, Faster-RCNN, RCNN, RetinaNet, and You Look Only Once (YOLO) and YOLO series are considered as milestones in the development of object detection algorithms. The former two models used CNN based two-stage Detectors, while YOLO is one of the CNN based One-stage Detectors. Due to the advantages in addressing the problem of data imbalance, RCNN and YOLO are seen as milestones in the algorithm development of object detection, winning popularity especially in industries.
\cite{2022Road} made comparative analysis among Region-based Fully Convolutional Network (R-FCN, \cite{2015Fully}), Mask Region-based Convolutional Neural Networks (Mask R-CNN, \cite{8237584}), Single Shot Multi-Box Detector (SSD, \cite{10.1007/978-3-319-46448-0_2}), and YOLOv4 in object detection, finding that YOLOv4 outperform in accurately detecting difficult road targets under complex road scenarios and weather conditions in an identical testing environment.  \cite{2019Computer} and \cite{9304730} both used R-CNN models for object detection of accidents from surveillance viewpoints and achieved a high Detection Rate and a low False Alarm Rate. \cite{Wang_2019} used YOLOv3 as the base model for feature extraction of targets in video datasets, and carried out object detection of accidents including vehicle rollovers.

From the perspective of application of datasets, video datasets are very suitable for object detection tasks because of their high information content and direct expression compared to CSV and text datasets. As for the algorithm development, the current algorithm models for traffic accidents detection are relatively well developed, but there are bottlenecks in development due to the limitation of data sources. Therefore, the dataset provided in this paper is of great importance for future practical applications.

\section{Traffic accidents dataset (TAD)}
\label{sec:Traffic Accidents Dataset}

This section describes how we constructed the TAD and presents its properties through statistical analysis and comparison versus existing traffic accident datasets. Though constructing a traffic dataset is an arduous task, we endeavor to build a large-scale traffic accident dataset named TAD from surveillance perspective in various scenes. TAD includes serious traffic incidents caused by rain, vandalism or other factors, with a total of 333 videos covering 261 positive ones with traffic accidents.

In Section \ref{sec:Construction of Dataset}, we firstly describe the whole process of how TAD is developed, including the methods deployed in collecting data, extracting frames and generating labels set for tasks on classification and detection. In Section \ref{sec:Dataset statistics}, statistical characteristics of TAD in various perspectives are described in detail through examples and explanation. We also compare TAD with several mainstream traffic accident datasets, as described in Section \ref{sec:TAD and related datasets}.

\subsection{Construction of dataset}
\label{sec:Construction of Dataset}
TAD is a large-scale traffic accident dataset from surveillance perspective, including serious traffic incidents caused by rain, vandalism or other accidental factors. As for the recording time, accident data we provided includes both at the time of the accident and after the accident; from the perspective of collection sources, our accident data includes those downloaded from traffic video analysis platforms and those downloaded from mainstream video broadcasters such as Weibo Sina. From the perspective of accident types, there are collisions between vehicles, collisions between cars and pedestrians or cyclists who suddenly cross the road, collisions between cars and road obstacles or roadside fences, and rollovers caused by emergency braking and other factors. We chose the video data with more obvious accident characteristics from the visual point of view, and the statistical distribution results are shown in Table \ref{tab:table1}.

\subsubsection{Collecting data}
\label{sec:Collecting data}

\begin{figure}
    \begin{subfigure}{.5\textwidth}
      \centering
      \includegraphics[scale=0.75]{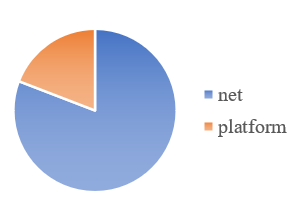}
      \caption{Data source distribution.}
      \label{fig:fig5a}
    \end{subfigure}
    \begin{subfigure}{.5\textwidth}
      \centering
      \includegraphics[scale=0.75]{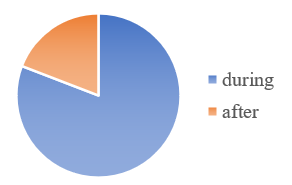}
      \caption{Time window distribution.}
      \label{fig:fig5b}
    \end{subfigure}
\caption{Overview distribution of TAD's data source.}
\label{fig:fig5}
\end{figure}

\textbf{Data source.} To ensure the quality of the dataset, we have edited the video data, keeping only the video time windows are located before, during and after a traffic accident. We used several sources to collect different categories of traffic accidents, and will continue to extend the data in the future. Since the preset position of surveillance cameras is diverse in different time, it is hard to capture continuous video frames covering the complete process of an accident. Therefore, images acquired from our traffic video analysis platform mainly covers time windows after an accident happens. Images collected from Internet video mostly provides time windows before and after an accident. In order to acquire relative instances as much as possible, we use several search terms for retrieval, including “vehicle wreck”, “road accidents”, “freeway accidents”, “non-motor vehicle accidents”, etc. Videos including blurred images, pranks, minor cuts, and minor rear-end accidents are not considered in the selection. Therefore, data finally retained from the two sources generally contains traffic accidents which are serious and recorded completely.
Accordingly, the ratio between data from Internet and our platform is 1:4 (Figure \ref{fig:fig5a}), sharing the same ratio of data before an accident happens and when an accident is happening (Figure \ref{fig:fig5b}). This is to ensure effective application on accident detection and warning as early as possible with a more comprehensive time period of accidents.

\textbf{Resolution.} The resolution of the video we acquired ranged from (862,530) to (1432,990). The video analysis platform is an AI application we developed which can automatically analyze traffic incidents in real-time from surveillance cameras. It could deliver abnormal events warnings such as pedestrians crossing the highway, vehicles parked on the side of road, traffic accidents in the format of captured images and video clips before and after the reported moment. The playback videos are also available to be downloaded in higher definition through network video recoder (NVR), which gives us access to higher-resolution data. Data collected from this source usually consists of 2-3 video clips per accident scenes with both near and far views. The resolution of data collected from Internet video platforms such as Weibo is rather lower. However, variation in resolution does not significantly affect the visual analysis towards traffic accidents since we use human-annotation in the labelling tasks.

\subsubsection{Cleaning data}
Videos we provided in our dataset have different memory size varied from less than 1M to over 10M. In order to catch the exact moment when the accident happened, we applied two methods to extract images from videos.

In Table \ref{tab:table1}, we display a summary of the datasets distributions in number of videos divided in different size. It is worth noting that we employ different frame gaps to extract images according to different size of video. Videos equal to or smaller than 5M accounts for 92 $\%$ in total while the proportion of videos equal to or smaller than 1M is 35 $\%$. This means videos included larger than 5M, or even 10M is less, as they are mostly a collection of several clips of traffic events. This step aims toward balancing image quantity between different size of videos, in order to capture the moment when a traffic accident happens and present as rich amount of data as possible.
\begin{table}
    \centering
    \begin{tabular}{cccc}
    \toprule
         Size& Number& Average Time (s) & Average Resolution (px)\\
    \midrule
        $<1M$ & 92 & 10 & (862,530)\\
        $>=1M\&\&<5M$ &	149 & 20 & (1382, 883)\\
$>=5M\&\&<10M$ & 16 & 40 & (1490,923))\\
$>=10M$ &	4 & 60 & (1432,990)\\
    \bottomrule
    \end{tabular}
    \caption{A summary of our TAD videos in the number of different video size.}
    \label{tab:table1}
\end{table}

\subsubsection{Annotating data}
Our dataset has a total of two levels of annotation, including video-level classification labels and image-level rectangular annotation boxes. In the first stage, our goal is to divide data into two sets including the presence and absence of incidents. Each shot is a series of consecutive video clips containing the trajectory of a moving target.

Images obtained from the same video clip would be divided into two candidate sets depending on whether there existing targets in each frame. For images spots before an accident happens, they are verified as the “No accident” category. For images demonstrating the on-the-spot scene of an accident, there are two issues to be addressed. First, due to lower resolution, some images extracted could not demonstrate distinctive features of the process when a traffic accident happens. Second, some images, though in high resolution, could not be distinguished with normal traffic scene, such as one car passing by another or normal masking in traffic. Considering these issues, the solution taken is to filter the labels and only keeps differentiated ones annotated in the “Accident” category.

\begin{figure*}
    \centering
    \includegraphics[scale =0.6]{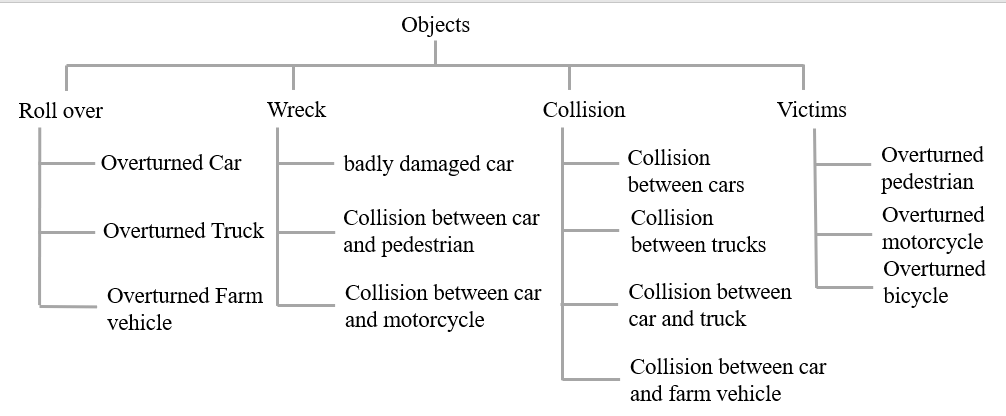}
    \caption{TAD classes.}
    \label{fig:tree}
\end{figure*}

In order to improve the quality of the dataset, we hire five professional workers to perform the labeling task, annotating the location and type in a box shape. Thereafter, the results of all annotators to make a more pertinent and stable data annotation. Finally, we combine the results of each worker with a second check to unify the annotation level and verify the quality. Therefore, TAD is constructed with multiple dimensions in terms of variety, clarity, and annotation. Positive labels used to indicate “Accident” are “collision”, “wreck”, “roll over” and “victims”, as illustrated in Figure \ref{fig:tree}.

\textbf{collision}
Collisions among motorized vehicles, such as collision between two cars.

\textbf{wreck}

-	Collisions between motorized and non-motorized vehicles, such as collision between a car and a motorcycle.

-	Collisions between pedestrian and motorized vehicles.

-	Injured motorized vehicles such as trucks and cars.

\textbf{roll over}
Rollovers.

\textbf{victims} 
Injured people as well as broken non-motorized vehicles such as bicycles.

\subsection{Dataset statistics}
\label{sec:Dataset statistics}

This section present results concluded from statistical analysis, showing the distribution of data upon different perspectives. This aims to give a clear illustration of TAD.

\begin{figure*}[ht]
  \begin{subfigure}{.5\textwidth}
    \centering
    \includegraphics[scale=0.8]{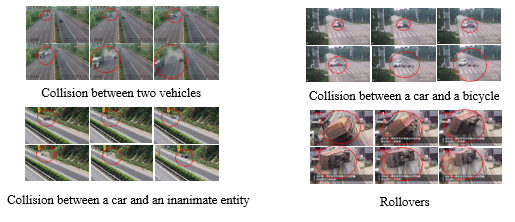}
    \caption{A snapshot of different kinds of traffic accidents in TAD.}
    \label{fig:accidents snapshot}
  \end{subfigure}
  \begin{subfigure}{.5\textwidth}
    \centering
    \includegraphics[scale =0.6]{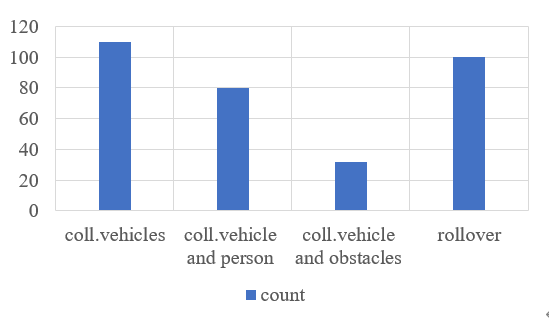}
    \caption{Shots distribution in different accident scenes of TAD.}
    \label{fig: accident distribution}
  \end{subfigure} 
  \caption{Different kinds of traffic accidents in TAD.}
  \label{fig:Different accidents in TAD }
\end{figure*}

\textbf{Type.} There are a total of four types of accidents in our dataset including collisions between multiple vehicles, collisions between vehicles and bicycle/motorcycle, collisions between vehicles and inanimate entity and rollovers. Figure \ref{fig:accidents snapshot} shows image sequences of the whole process of each four typical accident types. Each image sequence is placed in order and red dashed lines are marked for each image where the accident occurs.

Since some videos are a video collection integrated with several clips about one accident, we split the video into several individual clips when plotting the distribution. Instead of integrated videos, we count the number of each accident type at the level of individual video clips as depicted in Figure \ref{fig: accident distribution}.

\begin{figure*}[ht]
  \begin{subfigure}{.5\textwidth}
    \centering
    \includegraphics[scale=0.4]{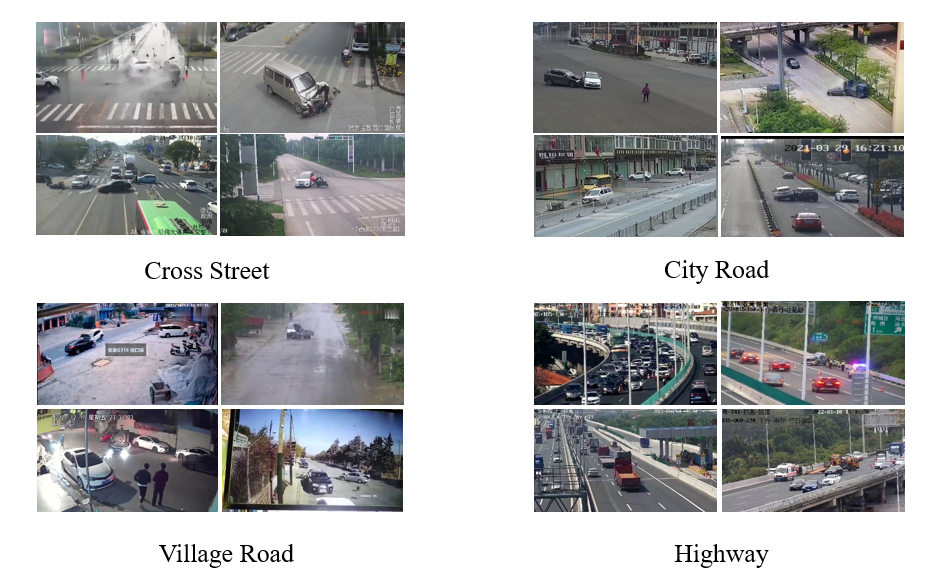}
    \caption{A snapshot of different scenes of TAD. }
    \label{fig:Scenes snapshot}
  \end{subfigure}
  \begin{subfigure}{.5\textwidth}
    \centering
    \includegraphics[scale =0.6]{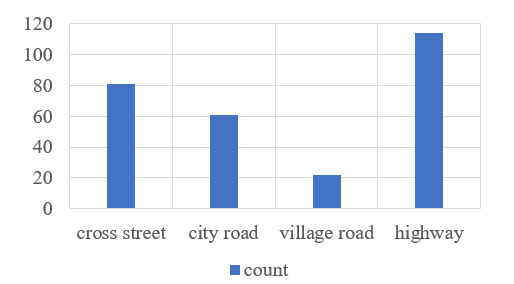}
    \caption{Shots distribution in different accident scenes of TAD.}
    \label{fig:Shots distribution}
  \end{subfigure} 
  \caption{Different scenes of traffic accidents in TAD.}
  \label{fig:different accident scenes of TAD}
\end{figure*}

\textbf{Accidents scenes.} There are totally four scenarios of accidents in TAD including junctions, urban roads, village roads, and highways together with expressways. Figure \ref{fig:Scenes snapshot} illustrates samples for each scene with randomly selected images. It is worth noting that even the same scenario contains different monitoring angle, enriching the visual information of each accident type.

The number of shots counted for each four scenes of TAD shows that data collected from “highway” scenes account for the largest among the four. This means that, TAD is a visual dataset in accident scenes, specifically from the monitoring view of highway (See Figure \ref{fig:Shots distribution}). This addresses the challenging problem facing visual recognition study where open-sourced dataset of traffic accidents under highway surveillance is rather rare. The proportion of the other three types is less, referring to “junction”, “urban road”, and “village road” respectively.

Generally, junctions are scenes where accidents are more likely to happen. Obviously, this is mainly due to higher traffic flow and probability of abnormal activities such as red-light running in junctions. Accidents happens in highways comprises the highest proportion of among the four scenes. Mainly providing highway accident data, TAD also offers images and visions where accidents happen in other type of scenes. This can aid to deeper exploration in the pattern and significant characteristics of accident occurrence, improving the efficiency in accident recognition.

\begin{figure}
    \centering
    \includegraphics[scale =0.6]{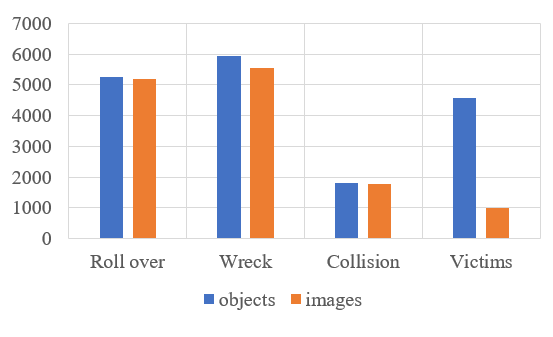}
    \caption{Number of objects and images per label. }
    \label{fig:Number of objects and images per label }
\end{figure}

\textbf{Scale.} TAD aims to provide a wide coverage of the accidents happening in the real world. There are 261 videos, with 278 video clips and totally provided in TAD. Figure \ref{fig:Number of objects and images per label } shows the histogram distribution of the number of images as well as objects per label.

\begin{table}
    \centering
    \begin{tabular}{cccccccc}
    \toprule
         Dataset& Total duration& Average \#frames & \#positives/\#all& C & H & S & T\\
    \midrule
       \textbf{TAD} & 1.2hours & 896 & 261/333 &Y&Y&Y&Y\\
       UCF-Crime &	- & - & 151/1900&Y&N&N&Y\\
       CADP & 5.2hours & 366 & 1416/1416 &Y&N&Y&Y\\
       DAD &2.4hours & 100 & 630/1730 & N&N&Y&Y\\
       HTA & - & - & - &N & Y &N&Y\\
    \bottomrule
    \end{tabular}
    \caption{Comparison results of TAD with mainstream anomaly detection datasets. \textbf{C}: CCTV footage, \textbf{H}:highway scene, \textbf{S}:spatial annotation,\textbf{T}:temporal annotation. }
    \label{tab:table2}
\end{table}

\subsection{TAD and related datasets}
\label{sec:TAD and related datasets}
Next, we present the properties of the traffic accident dataset (TAD) in comparison versus other popular traffic incidents datasets. The main differences are summarized in Table \ref{tab:table2}.

Traffic accidents are often considered as one category of traffic anomaly detection and open-sourced video datasets related to traffic accidents are limited. Therefore, dataset comparison in this section targets traffic video datasets commonly applied in traffic anomaly detection. 

\textbf{Third-person perspective.} HTA and DAD provide videos of accidents captured by dashcam mounted on driving vehicles, providing the first-person perspective. Since More information on surrounding conditions with a wider view is provided in videos from CCTV footage, such as traffic flow, vehicles driving in front of and after the target car. UCF-Crime, CADP and our
datasets provide surveillance videos, which is critical
to improving intelligent traffic system.

\textbf{Accident scenes.} Videos we collected reflect real accidents mainly happens in highways as shown in Figure 7.
Since the speed of vehicles driving on highways is often higher than that of urban roads or streets, accidents caused in highways often happens in very short time, making it difficult to collect positive samples. UCF-Crime, CADP and DAD, though large in scale, lack scenes recorded in highways. HTA and Ours record highway-level visual observation of traffic accidents. Other scenes “junction”, “urban road”, and “village road” generally seen are also included in HTA for comprehensive visual understanding as well as application.

\textbf{Annotation types.} Clearer annotation at the level of accident types is rarely seen in recent public traffic accident
datasets. CADP’s annotations are made upon temporal-spatial
level rather than accident types. UCF-Crime dataset concludes all types of accidents in one single category. So does
HTA. TAD is also temporal-spatial level. While our dataset  provide detailed annotation in describing accidents, including rollovers, and collision between vehicles, collision between vehicles and pedestrian or cyclist suddenly crossing, collision between a vehicle and a road obstacle or roadside fence, rollover caused by its own cause (emergency braking). These diverse visual information offers complementary resources for the vision development.

\textbf{Scale and length.} There are 331 videos with 261 posi-
801	tive samples with accidents included in our dataset.  
The total duration of TAD is 1.2hours in total. The longest video has 5.2hours and DAD has 2.4hours. But our datasets has the longest frames number in average. TAD is 896 frames per
805	video, which means more complete process provided in each accident.

\section{Traffic events benchmark}
\label{sec:Traffic Events Benchmark}

TAD randomly extracted 24,810 labeled images from 333 videos, 4 main accident types. Most analysis will be based on the current version. In this section we implement several experiments in order to test the performance of our dataset and provide a reference in application. Therefore, experiments are organized in image classification, video classification and object detection tasks. In each task, algorithms we selected are classical, mainstream and state-of-art. To better serve as a benchmark for visual algorithms, TAD is split into train, validation and test sets.

\subsection{Image classification}
\label{sec:Image classification}

Experiments carried out in this section aim to verify classification algorithms on TAD to check their performance in detecting traffic accidents. Since TAD is collected from two sources varied from resolution and the number of accident types, we divide TAD into two categories referring to “TrafficAccident-net” and “TrafficAccident-platform”, which indicate they are selected from internet video sharing platforms and the traffic analysis platform we employed in industrial projects respectively. Similarly, dataset “TrafficAccident-all” indicate the merged version from above two sources.

In order to better present TAD’s performance, we design all experiments based on comparison within three datasets, namely “TrafficAccident-net”, “TrafficAccident-all” and RoadAccident (\cite{2018Real}). On the one hand, comparison between “TrafficAccident-net” and “TrafficAccident-all” aims to clearly present the improvement of performance thanks to data collected from traffic analysis platform. The two training sets each separate 10$\%$ of data as validation set and the rest 90$\%$ is used for training. On the other hand, we also choose RoadAccident, a dataset closer to ours, as the control group to be tested on our classification experiments for accidents detection. All the three datasets share the same test data, in toal 1490 images randomly selected 30 scenes downloaded from traffic analysis platform. The ratio of positive and negative data is kept 1:1 in the test.

Our image classification experiments are based on ResNet50 (\cite{2016Deep}) as its backbone in Convolutional Architecture for Fast Feature Embedding (Caffe) deep learning frameworks running on one GPU of NVIDIA GeForce RTX2080Ti. We divide whole images into two categories: “Accident” and “Normal”, which refer to scenarios with and without accidents respectively. In Figure \ref{fig:Annotation samples}, the above three lines of images for each side illustrate the complete process of an accident, while the fourth line are those with a single shot of an accident. The left refers to the positive samples with distinctive features of accidents, while the right shows negative ones without accidents.

The comparison results of image classification task within the three datasets evaluated by Recall, Precision and F1-score in Table \ref{tab:table3}. RoadAccident tops in Precision because of its singularity in accident types which helps to achieve model convergence while declines its recall and F1-score. This indicates its incapability in detecting the majority of accidents. With richer accident information provided, TrafficAccident-net (Part TAD) and TrafficAccident-all (TAD), though receive lower precision, are superior in Recall and F1-score. With TrafficAccident-net (Part TAD) leading in Recall, we could find that data of TAD are conducive to detect accidents compared to RoadAccident. Besides, The inclusion of data with higher resolution collected from traffic analysis platform contributes to feature learning, making it easier for the model to capture characteristics of accidents. Therefore, TrafficAccident-all (TAD) performs better than TrafficAccident-net (Part TAD) in Precision and F1-score, which demonstrates excellency of larger scale and higher quality of TAD.

\begin{figure*}
    \centering
    \includegraphics[scale= 0.6]{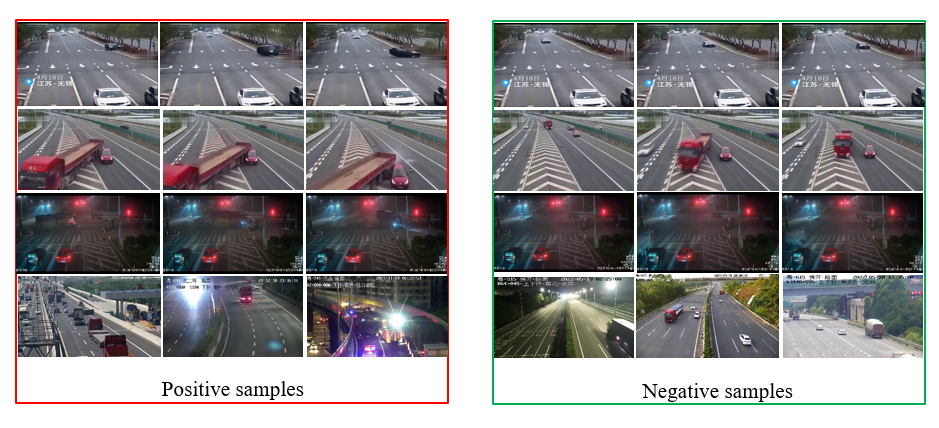}
    \caption{A snapshot of positive and negative samples for image classification.}
    \label{fig:img cls samples}
\end{figure*}

\begin{table*}
    \centering
    \begin{tabular}{cccc}
    \toprule
      Dataset   & Recall & Precision&F1-score  \\
      \midrule
         TrafficAccident-net
 & 0.46 & 0.516 & 0.486\\
         TrafficAccident-all
&0.443&0.545&0.489\\
RoadAccident  & 0.04 & 0.6 & 0.075\\
         
    \bottomrule
    \end{tabular}
    \caption{Comparison results of image classification task in different datasets.}
    \label{tab:table3}
\end{table*}

\begin{table}
  \centering
  \begin{tabular}{cccc}
 
    \toprule
     Model	&Recall	&Precision	&F1-score \\
    \midrule
    R3D	&0.895	&0.548	&0.679 \\
    Slowfast-ResNet50 &1.000	&0.515	&0.680 \\
    \bottomrule
  \end{tabular}
  \caption{Comparison results of TAD in different video classification algorithm.}
  \label{tab:table4}
\end{table}

\subsection{Video classification}
\label{sec:Video classification}
\begin{figure*}[ht]
  \begin{subfigure}{\textwidth}
    \centering
    \includegraphics[scale=0.75]{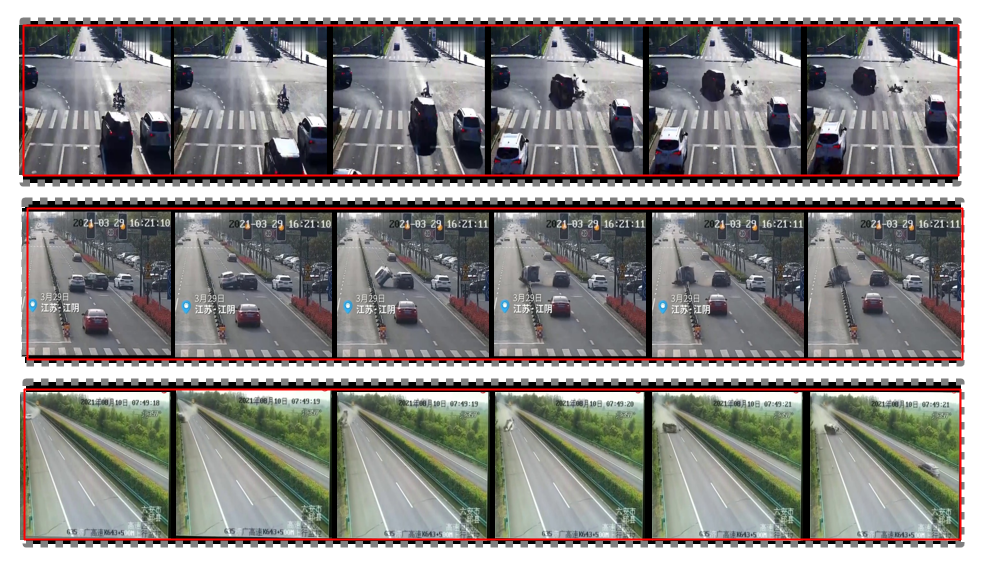}
    \caption{Positive samples.}
    \label{fig:fig14a}
   \end{subfigure}

  \begin{subfigure}{\textwidth}
    \centering   
    \includegraphics[scale=0.75]{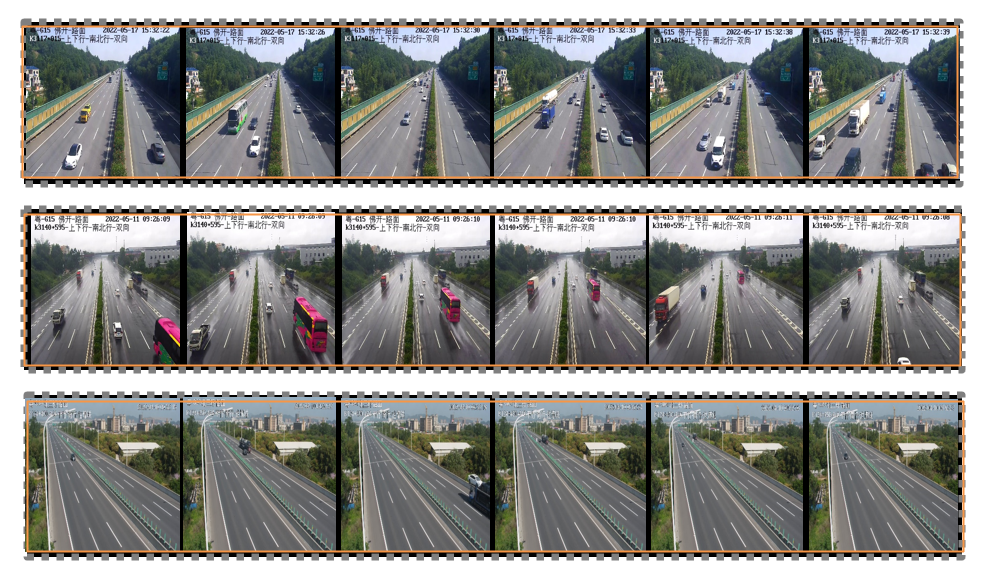}
    \caption{Negative samples.}
    \label{fig:fig14b}
  \end{subfigure}
  \caption{Positive and negative samples for video classification.}
  \label{fig:fig14}
\end{figure*}

Video classification is almost the same as the task of image classification except the data format. Similarly, video classification task is designed to differentiate videos into two types referring “Accident” and “Normal” based on mainstream video classification models. In this section, 261 videos are used as positive samples and 72 videos without accidents are taken as negative ones. We construct the training set and validation set with the ratio of 9:1, while the test sat remains the same video source utilized in \ref{sec:Image classification}. As shown in Figure \ref{fig:fig14a}, positive samples with consecutive sequences used in video classification are framed with red lines one (Figure \ref{fig:fig14a}) with negatives ones framed in orange (Figure \ref{fig:fig14b}).

Models employed are R3D (\cite{20133D}) and SlowFast (\cite{2019SlowFast}), which are both classical algorithms utilized in video classification. Using ResNet50 as its backbone. Slowfast owns less parameters as well as computational cost than R3D. Both networks use the same configuration of parameters with clip len setting 16 and frame interval setting 4 trained in Pytorch deep learning framework. The original size of images is randomly selected from 128 to 171. Later we randomly crop the images into (112,112) size square covering the target object. Finally, the input to the Neural Network are images sized in (16,112,112).

Models are all trained from scratch for 200 epochs with initial learning rate set at le-3. Evaluation results of “TrafficAccident-all”(TAD) trained on R3D and Slowfast-ResNet50 are compared in Table \ref{tab:table4}. Here we employ the same evaluation metrics with image classification. As described in Table, R3D is 3$\%$ higher in Precision and almost 1$\%$ lower in Recall, compared to Slowfast-ResNet50. Both models share close performance in F1-score. nearly 68$\%$.


\subsection{Object detection}

Experiments in this section are designed based on images in order to verify capability of spotting and detecting accidents of mainstream object detection models. We choose YOLOv5 \cite{2016You} with version 4.0 in Pytorch as our training model, with different backbones. “YOLOv5x”, and “YOLOv5m” are all employed for comprehensive evaluation on accuracy and speed of YOLOv5.

\begin{figure*}
    \centering
    \includegraphics[scale = 0.6]{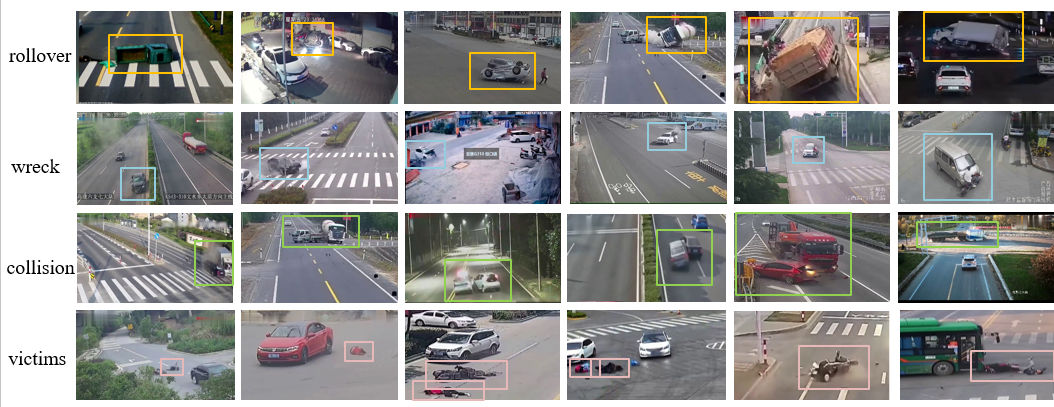}
    \caption{Annotation samples for object detection.}
    \label{fig:Annotation samples}
\end{figure*}

The two versions of TAD, both “TrafficAccident-net” (Part TAD) and “TrafficAccident-all” (TAD) are trained with the same test data and logic applied in \ref{sec:Image classification} and \ref{sec:Video classification}. Dataset for object detection are annotated in Labelme into four labels, referring “roll over”, “wreck” , “collision” and “victims”. Figure \ref{fig:Annotation samples} shows some annotation samples applied in object detection. Each line consists of 6 shots with targets labeled. It’s worth noting that target objects are bounded slightly bigger than its original size. Specifically, each bounding box is expanded 1/3 times from the minimum external rectangle box of target area. This intends to better capture background features apart from the target in the area where accidents happen, improving detection performance of recognizing accidents.


\begin{table}
    \centering
    \begin{tabular}{cccc}
    \toprule
      Dataset   &AP&mAP(0.5)&AR  \\
    \midrule  
      TrafficAccident-all &0.525 & 0.422 & 0.419\\
      TrafficAccident-net & 0.414 & 0.39 &0.333\\
    \bottomrule  
    \end{tabular}
    \caption{Detection performance of TAD with object detection task (score>0.4) among different subsets.}
    \label{tab:table5}
\end{table}

\begin{table}
    \centering
    \begin{tabular}{cccc}
    \toprule
      Dataset   &AP&mAP(0.5)&AR  \\
     \midrule
     YOLOv5x-960-single &0.525 & 0.422 & 0.419\\
     YOLOv5m-960-single & 0.518 & 0.39 &0.319\\
      \bottomrule
    \end{tabular}
    \caption{Detection performance of TAD with object detection task (score>0.4) among different backbones.}
    \label{tab:table6}
\end{table}

Experiments are conducted on two GPUs of NVIDIA GeForce RTX2080Ti in Ubuntu system. Evaluation metrics for objection detection are AR, AP and MAP(0.5) while results for tests on single image are measured by Recall, Precision and F1-score to present comparison with image classification. 

Table \ref{tab:table5} shows the comparative analysis of training results between TrafficAccident-all(TAD) and TrafficAccident-net(Part TAD) on the test data. TrafficAccident-all outperforms TrafficAccident-net though with the same training configuration of parameters and environment. The two datasets show a significant to learning features of real-life accidents. There is slight difference in mAP within object categories, which is possibly owing to difference in the sample size distribution or feature learning capability.

In order to compare backbones with different parameters and complexity on test data, another experiment is designed between YOLOv5x, the larger one and YOLOv5m, as illustrated in Table \ref{tab:table6}. Difference between the two models is larger in AR and smaller in AP as well as precision, which indicates more complex network facilitates the detection of accidents, thus effectively reducing the false negative rate(FNR).

On the other hand, we also organize experiments on images to judge the occurrence of accidents by setting 0.4 as the threshold confidence. That is, images predicted with scores higher than 0.4 would be classified as positive samples for accidents detection. Evaluation metrics applied for this task are also Recall, Precision and F1-score, as shown in Table \ref{tab:table7}. It’s noticed that two models present almost identical performance in the binary classification of accidents detection. Therefore, considering the cost and effect of model complexity and detection accuracy, YOLOv5m is selected as the backbone for the subsequent detection experiments.

\begin{table}
    \centering
    \begin{tabular}{cccc}
    \toprule
      Dataset   & Recall & Precision & F1-score  \\
    \midrule  
      YOLOv5x-960-single
 &0.997 & 0.487 & 0.654\\
      YOLOv5m-960-single
 & 0.995 & 0.486 &0.655\\
    \bottomrule  
    \end{tabular}
    \caption{Classification performance of TAD with object detection task (score>0.4) among different backbones.}
    \label{tab:table7}
\end{table}

\begin{figure*}
    \centering
    \includegraphics[scale=0.6]{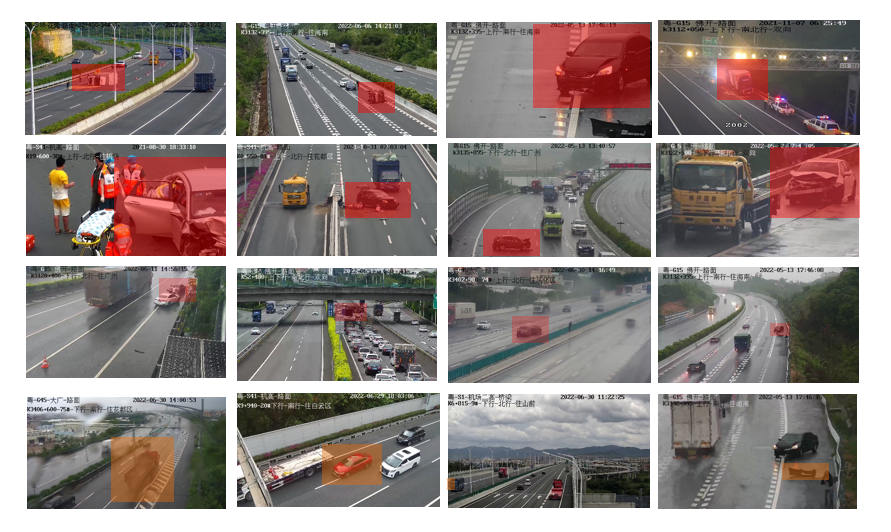}
    \caption{Visualization results for the test data.}
    \label{fig:fig13}
\end{figure*}

Figure \ref{fig:fig13} visualizes test results of in our object detection experiments. Areas colored in red refers to targets receiving accurate detection in first three rows, while the yellow colored ones, located in the bottom, indicate wrong predictions. Specifically, it’s worth noting that the top two rows are darker in red for higher confidence in detecting “wreck” and “roll over” compared to the third row. Lower confidence mainly due to smaller size of distant targets, making it harder for capturing visual features, which could be improved by model iterations through addition of more negative samples.

\subsection{Experiments summary}
The above three experiments focus on traffic accident prediction methods based on various vision tasks. 

The classification and object detection tasks we explore are image level, learning and utilizing two-dimensional visual features of images. Results show that classification algorithms mainly learn and capture the global features of images, focusing more on the overall regions of images. In contrast, guided by the supervised learning logic, detection algorithms are more refined as they focus on local regions while mining the information of accident types and locations. As shown in Table \ref{tab:table8}, the recall rate of object detection is higher than those of classification, indicating that fine-grained regions of accidents are more beneficial to comprehensive detection on accident types. In terms of accuracy, the detection algorithm is currently lower than classification, possibly due to the difference caused in prediction on each category. Leading high false alarm rates, the two types of accidents, “collision” and “victims”, pull down the overall accident prediction accuracy. In fact, the best way to discern the occurrence of an accident is to analyze it by the whole process of its occurrence. Therefore, we perform an accident detection detection from the complete process of accidents in final part of the entire experiments by using the average classification results of a series of video frames for different tasks.

\begin{table*}
    \centering
    \begin{tabular}{ccccc}
    \toprule
      Task   &Model&Recall&Precision&F1-score  \\
      \midrule
         Image classification& ResNet50 &0.579&0.595&0.454\\
         Image detection& YOLOv5m &0.997&0.487&0.654\\
         
    \bottomrule
    \end{tabular}
    \caption{Image-level classification performance of TAD among different vision tasks.}
    \label{tab:table8}
\end{table*}

\begin{table*}
    \centering
    \begin{tabular}{ccccc}
    \toprule
      Task   &Model&Recall&Precision&F1-score  \\
      \midrule
         Image classification& ResNet50 &1.00&0.5&0.666\\
         Image detection& YOLOv5m &0.261&1.00&0.414\\
         Video classification& SlowFast-ResNet50&1.00&0.515&0.680\\
    \bottomrule
    \end{tabular}
    \caption{Video-level classification performance of TAD among different vision tasks.}
    \label{tab:table9}
\end{table*}

Table \ref{tab:table9} shows the prediction results of the three algorithms on the test set. With the same classification algorithm, the video classification model carries higher accuracy than the image classification model. This is because the model for video classification also focus on the sequential information of the time dimension in addition to capturing the two-dimensional features of the image. Therefore, it’s more accurate in discriminating whether an accident has occurred.

The recall rate of the image level target detection algorithm decreases significantly when compared to the classification algorithm, considering the results of all frames together, indicating that the target detection model is very unstable in discriminating results between video frames, resulting in true accidents missed in the final recognition results. On the contrary, only a consistent and stable output will confirm that it is an accident and the detection accuracy will be very accurate. This result also shows that traffic accident prediction needs to combine information about the temporal dimension between consecutive video frames for a more comprehensive and accurate accident discrimination.

\section{Discussion}
\label{sec:Discussion}

 A new traffic accidents dataset, covering abundant accident scenes, from CCTV footage named TAD was introduced in this work. In which, a wider collection of accident types was collected, annotated and constructed to press ahead the development of visual recognition of accidents happened on roads. Moreover, TAD retains series of accident types with distinctive features. It includes rollovers and collisions not only caused by multiple vehicles, but also between vehicles, bicycles, pedestrians as well as inanimate entities. Statistical comparison and algorithmic analysis both indicate that TAD performs advantages in information dimension. We envision two main possible aspects for application as follows:
 
\emph{A benchmark dataset.} TAD aims to play the role as a benchmark for extensive research on visual application. We believe that accident datasets with high quality, specificity and large scale will enable the advancement in object detection and visual classification tasks.

\emph{Accident vision research.} TAD offers annotation for multiple types of accidents from CCTV footage. This caters to the desperate need for open-sourced datasets on traffic research.

\section{Future work}
\label{sec:Future Work}

Here we list no-exhaustive respects to be improved for further application and advancement.
we intend to include more objects in our dataset. More specifically, we propose to refine the annotation of vehicles in categories such as car and truck. This would add to more visual information provided by a certain accident video for future research. More scenes of accidents with further variation in weather and roads are expected to concluded into extended version of TAD. We plan to enrich the categories of accidents. For examples, we are about to divide the label “collision” into several subcategories, such as head-on collision, rear-end collision, side-impact collision.

\clearpage

\end{document}